\documentclass{article}

\usepackage[final]{neurips_2022}

\usepackage[utf8]{inputenc} 
\usepackage[T1]{fontenc}    
\usepackage{hyperref}       
\usepackage{url}            
\usepackage{booktabs}       
\usepackage{amsfonts}       
\usepackage{nicefrac}       
\usepackage{microtype}      
\usepackage{xcolor}         
\usepackage{graphicx}
\usepackage{bm}
\usepackage{natbib}

\usepackage{wrapfig}
\usepackage{amsmath}
\usepackage{hyperref}
\usepackage{multirow}
\usepackage{booktabs}
\usepackage{tabularx}

\newcommand{\myparagraph}[1]{\vspace{3pt}\noindent{\bf #1}}

\title{CosSGD: Communication-Efficient Federated Learning with a Simple Cosine-Based Quantization}

\author{
   Yang He$^{1}$\thanks{Work done before joining Amazon.}\ , \  Hui-Po Wang$^{1}$, \ Maximilian Zenk$^{2}$, \ Mario Fritz$^{1}$ \\
   $^{1}$ CISPA Helmholtz Center for Information Security\\
    $^{2}$ Division of Medical Image Computing, German Cancer Research Center (DKFZ) \\
\texttt{yanhea@amazon.com, \{hui.wang,fritz\}@cispa.saarland,} \\ \texttt{m.zenk@dkfz-heidelberg.de} 
}

\begin{document}

\maketitle

\begin{abstract}
  Federated learning is a promising framework to mitigate data privacy and computation concerns. However, the communication cost between the server and clients has become the major bottleneck for successful deployment. 
Despite notable progress in gradient compression, the existing quantization methods require further improvement when low-bits compression is applied, especially the overall systems often degenerate a lot when quantization are applied in double directions to compress model weights and gradients.
In this work, we propose a simple cosine-based nonlinear quantization and achieve impressive results in compressing round-trip communication costs. We are not only able to compress model weights and gradients at higher ratios than previous methods, but also achieve competing model performance at the same time.
Further, our approach is highly suitable for federated learning problems since it has low computational complexity and requires only a little additional data to recover the compressed information.
Extensive experiments have been conducted on image classification and brain tumor semantic segmentation using the CIFAR-10, and BraTS datasets where we show state-of-the-art effectiveness and impressive communication efficiency. 
\end{abstract}

\section{Introduction}
\begin{wrapfigure}{r}{0.45\textwidth}
\vspace{-40pt}
\begin{center}
\includegraphics[width=\linewidth]{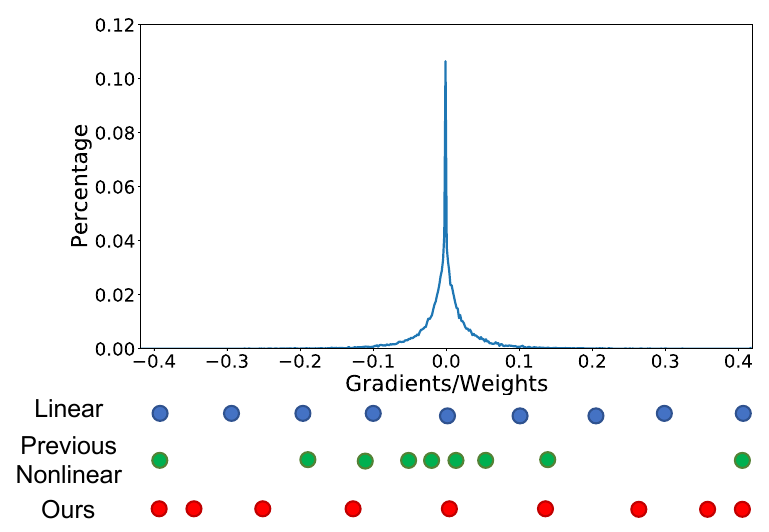}
   \end{center}
   \vspace{-0.5cm}
   \caption{Quantization intervals utilized in the proposed method and prior works.} 
   \vspace{-0.7cm}
\label{fig:teaser}
\end{wrapfigure}

Recently, Federated Learning (FL)~\cite{mcmahan2017fl_google} has emerged as a prominent approach to meet the heavy demands on data privacy and computation resources. It allows a vast number of edge devices, called workers, to train a model collaboratively without sharing private data.
Due to the effectiveness of FL, it has been widely adopted in many practical applications, such as keyboard prediction~\cite{hard2018keyboard_prediction_fl, yang2018applied_keyboard} and privacy-sensitive services~\cite{sheller2018multi_institutional,sheller2018multi_brain_fl,pokhrel2020fl_block_chain}.

However, when training FL models, the model weights and gradients are transmitted many times between the server and workers. As the model complexity grows, the communication cost inevitably becomes the major bottleneck of FL systems. To alleviate the issue, many works have made efforts to apply quantization on the model weights~\cite{yu2019double_quantization,tang2019doublesqueeze} or gradients~\cite{wen2017terngrad,alistarh2017qsgd} to reduce the message size of model updates.
Although the above works have achieved great progress, they often fail to train a model with the similar model performance to float based non-compressed training, when heavy compression is deployed. Besides, we observe low-bits quantization may deteriorate other widely used techniques (e.g., gradient sparsification~\cite{aji2017sparse_top_k,strom2015scalable_top_k}), which limits the further cost reduction.

In this work, we propose a novel quantization method based on cosine function, which is rather simply but effective to quantize model weights and gradients for round-trip compression. Inspired by the key observation that gradients and model weights with larger magnitudes are more informative than the others~\cite{han2015deep_compression,lin2018deep_gradient_compression,aji2017sparse_top_k}, we leverage the non-linearity of cosine functions for minimizing the quantization errors of extreme values, whereas most prior works~\cite{alistarh2017qsgd,han2015deep_compression} aim for an opposite goal as shown in Fig.~\ref{fig:teaser}. 

In addition, our method has low time/space complexity, requires \textit{no} additional training, and introduces only little additional cost, which makes our approach further favorable to federated learning systems since the edge devices are often equipped with limited hardware resources.

We conduct extensive experiments under various scenarios on CIFAR-10 for image classification as well as BraTS 2019 dataset for 3D volume segmentation. We depict our contributions as follows: (a) As the key contribution, we propose a \textit{simple-yet-effective} cosine-based quantization scheme. It better preserves model performance, has low time/space complexity and requires no extra training. It is thus applicable to a variety of distributed learning problems.
(b) Our proposed quantization can be applied to compress model weights and gradients to reduce overall communication costs in federated learning.
(c) Our method is flexible and compatible to many existing useful techniques such as gradient sparsification. By combining our quantization with other techniques, our system is able to obtain the similar model performance to the float based non-compressed training at very high compression ratios, while other quantization methods are hard to achieve.

\section{Related work}

\vspace{-0.2cm}
\subsection{Federated and distributed learning}
Federated learning~\cite{yang2019federated_concept,lim2020federated_survey,kairouz2019open_fl,li2020federated_future} is a specialized distributed learning framework that allows a vast number of edge devices to jointly train a model while keeping private training data on devices. Federated Averaging (FedAvg)~\cite{mcmahan2017fl_google} is one of the most classical optimization methods. 
Recently, FL improves with adaptive optimization. Reddi~et al.~\cite{reddi2020adaptive_fl} applies an optimizer on the server with momentum terms to update the model with aggregated gradients, which shows faster convergence and higher performance in many tasks. Mime~\cite{karimireddy2020mime} uses server-level momentum in each client to mitigate the client-drift caused by the heterogeneous data across clients and reach higher performance. FedCD~\cite{kopparapu2020fedcd} dynamically groups the local workers with similar data distributions to improve the model performance on Non-IID data.

\vspace{-0.2cm}
\subsection{Gradient compression}
We discuss gradient quantization~\cite{alistarh2017qsgd,fu2020tinyscript,han2015deep_compression,konevcny2016fl_communication,seide20141} and sparsification~\cite{konevcny2016fl_communication,aji2017sparse_top_k, strom2015scalable_top_k, alistarh2018convergence_top_k,lin2018deep_gradient_compression} in this subsection.
Quantization aims to compress the float gradients to low-bits representations.
Due to biased approximation, systems may lose performance after quantization. To overcome this, two parallel works, TernGrad~\cite{wen2017terngrad} and QSGD~\cite{alistarh2017qsgd}, utilize probabilistic unbiased quantization with correct expectations for quantized gradients.
In addition, quantization errors can be reduced by rotations with random Hadamard matrices, which need extra computation~\cite{suresh2017distributed_mean_hadamard}. 
Besides, signSGD~\cite{bernstein2018signsgd} shows that the gradient signs -- represented with only 1-bit -- can be used for optimizing non-convex problems. Lastly, nonlinear quantization methods~\cite{han2015deep_compression, fu2020tinyscript} also study producing fewer quantization errors for the densely distributed data. 
Meanwhile, communication efficiency is vastly improved by combining with sparsification, which aims to 
return parts of gradients. The main strategies include random locations~\cite{konevcny2016fl_communication} and top-K locations~\cite{aji2017sparse_top_k, strom2015scalable_top_k, alistarh2018convergence_top_k,lin2018deep_gradient_compression}.

Different from prior work, we design a nonlinear quantization with larger quantization intervals for the less important values and more precise intervals for the values playing a more important role for training.

\vspace{-0.2cm}
\subsection{Reducing overall communication costs}
So far, there are only few works on reducing overall communication costs. Existing works often approach the problem by simplifying the model architectures communicated between the server and workers. For example, Federated Dropout~\cite{caldas2018expanding} and Adaptive Federated Dropout~\cite{bouacida2020adaptive}, motivated by the idea of Dropout, stochastically construct slimmed sub-nets; Liang~et al.~\cite{liang2020think} divide the overall network into local feature extractors and a global classifier. The feature extractors are kept on local devices while only the classifier part is communicated. In addition to federated learning, quantizing model weights and gradients have been explored in general distributed learning~\cite{tang2019doublesqueeze,yu2019double_quantization} with linear quantization~\cite{alistarh2017qsgd} to reduce the communication costs for both directions.

As a compensation, we propose a more effective quantization than previous widely used linear quantization~\cite{alistarh2017qsgd,konevcny2016fl_communication} and other methods~\cite{han2015deep_compression,fu2020tinyscript,karimireddy2019error_feeback_signsgd,xie2020cser} for model weights and gradients, which is compatible with many existing frameworks.

\section{Nonlinear quantization with {\it \textbf{cosine}} function}
\label{sec:method}
In this work, we apply the quantization on model weights (server-to-client) and gradients (client-to-server) to reduce the overall round-trip communication costs.
To facilitate more effective compression, we propose a novel cosine-based quantization scheme, which achieves high compression ratios and is still manageable to train federated models.
In this section, we first describe our cosine-based quantization method and then discuss the properties of the proposed quantization. 

 For weights and gradients, they have similar properties that the values around 0 are much more than larger values. 
Besides, the principles for compressing them are similar in that we observe the larger values play a more important role in training, as pointed out by~\citep{han2015deep_compression}.

In federated learning or data-parallel optimization, the workers share the same model $M=(w_1,...,w_n)^T$ and update it by local data without explicit interactions between workers. After learning with local data, the gradients for the parameters $\nabla M$ are sent to the server. Accordingly, the model is updated by aggregating the gradients~\cite{mcmahan2017fl_google} from all the selected workers $\{\nabla M_i\}_{i=1}^{m}$ with the following formulation:

\vspace{-0.35cm}
\begin{equation}
    M^{t+1} = M^t - \eta_{s} \cdot \frac{\sum_{i=1}^{m} \nabla M_i\cdot N_{i}}{\sum_{i=1}^{m} N_i},
    \label{eq:model_update}
\end{equation}
where $t$ is the time step, $\{N_i\}_{i=1}^{m}$ are the numbers of training data on individual workers, and $\eta_{s}$ is the learning rate on the server.

To reduce the round-trip communication costs, we utilize quantization on the model weights from the previous step $M^t$ and gradients $\nabla M_i$ on each client.
The vector to quantize can be depicted by a set of variables or the angles of the vector to the standard coordinate.
We encode the angle information during quantization.
Let  $\textbf{v}$ be a vector, where $\textbf{v}=(v_1,...,v_n)^T$. The angle between the vector and $i^{-\text{th}}$ axis $\textbf{a}_i=(\underbrace{0,...,0}_{i-1},1,...,0)^T$
is computed by the inner product:
\vspace{-0.25cm}
\begin{equation}
    \cos(\theta_{i}) = \frac{\textbf{v}^{T}\textbf{a}_i}{\|\textbf{v}\|_2\cdot 1}=\frac{v_i}{\|\textbf{v}\|_2}.
\end{equation}
Accordingly, we have $\theta_{i}=\arccos(\frac{v_i}{\|\textbf{v}\|_2})$, where $\theta_{i}\in[0,\pi]$.

Different from previous linear-based uniform quantization~\cite{alistarh2017qsgd,konevcny2016fl_communication}, we quantize each value in the form of angles $\theta$ instead of the original values or the linearly transformed values~\cite{suresh2017distributed_mean_hadamard} in the Euclidean space. Specifically, in a normalized vector, there are very few or even no values close to 1 or -1, especially when the dimension of the vector is high. Therefore, we do not quantize the angle vector $\bm{\Theta}={(\theta_1,...,\theta_n)}^T$ on $[0,\pi]$; instead, we compute a bound $b_{\theta}=\min(\min(\Theta),1-\max(\Theta))$ and quantize $\bm{\Theta}$ on $[b_{\theta},\pi-b_{\theta}]$. The bound facilitates a more precise quantization in that it avoids the waste of quantization bins for the place where there is no value distributed. 

In addition to setting the exact $b_{\theta}$ for a vector, we also clip the top dimensions alternatively. Sometimes, there is one dimension dominating the gradient or weight vector. It will lead to a very large $b_{\theta}$, whereas most gradients (or weights) are small. Hence, the intensely dominating dimension prevents the others from making full use of the quantization space. To avoid this situation, we clip the top gradients or weights to get the bound $b_{\theta}$.
In the end, our quantization is a nonlinear operation that the space is partitioned unequally for the whole distribution.

We refer to the above vector quantization as $Q_v$, which is built on $Q_{\theta}$, a biased linear quantization on the angle space.
As pointed by~\cite{alistarh2017qsgd,wen2017terngrad}, the original quantization is biased
because $\mathbb{E}[Q_{\theta}(\bm{\Theta})]\neq\bm{\Theta}$. Accordingly, an unbiased quantization is achieved by a probabilistic procedure. Likewise, our cosine-based quantization can be extended with the probabilistic scheme, satisfying the unbiasedness. Formally, our unbiased $s$-bits quantization is defined as

\vspace{-0.25cm}
\begin{equation}\label{eq:unbiased_quan}
     Q_{\theta}(\bm{\Theta};b,s)=\left\{
\begin{array}{cl}
\lfloor v \rfloor         & \text{with} \quad 1-p \\
\lfloor v \rfloor + 1          & \text{otherwise}
\end{array} \right.,
\end{equation}
where $v=\frac{\bm{\Theta}-b}{\pi-2b}\times(2^s-1)$, $p=v-\lfloor v \rfloor$, and $\lfloor \cdot \rfloor$ is the round down operator. 
Finally, both our biased and unbiased methods return a quantized vector $Q_{\theta}(\bm{\Theta})$, the norm of the original gradient vector $\|\textbf{v}\|_2$, and the bound $b_{\theta}$. The gradients can be computed on the server by reversing the process.

\subsection{Quantization error bound analysis}
As depicted above, we quantize the angle vector $\bm{\Theta}$ over $[b_{\theta},\pi-b_{\theta}]$. Due to the symmetric property of the cosine function on $[b_{\theta},\pi-b_{\theta}]$, we focus on analyzing our biased quantization in the range of $[b_{\theta},\frac{\pi}{2})$, and the left holds the same rule on an opposite direction. The unbiased version can be estimated by taking the expectation form into considerations. 

The angle bound $b_{\theta}$ corresponds to a value (weights or gradients) $b_v$ (i.e., $v_i\in[-b_v, b_v]$), where $b_{\theta}=\arccos(\frac{b_v}{\|\textbf{v}\|_2})$. 
Let $q=\frac{\pi-2\cdot b_{\theta}}{2^s-1}$ be the width of angle quantization intervals, and then the quantization error of $v_i$ is bounded by

\vspace{-0.25cm}
\begin{equation} \label{eq:error_bound_cosine}
\begin{split}
|v_i-Q_v(v_i)| & \leq [\cos(q\cdot(k+\frac{1}{2})) - \cos(q\cdot(k+1))]\cdot\|\textbf{v}\|_2 \\
 & = 2\cdot\sin(q\cdot(k+\frac{3}{4}))\cdot\sin(q\cdot\frac{1}{4})\cdot\|\textbf{v}\|_2,
\end{split}
\end{equation}
where $k=\lfloor (\arccos(\frac{v_i}{\|\textbf{v}\|_2})-b_{\theta})/q \rfloor$, because $\cos(\theta-\epsilon)-\cos(\theta)<\cos(\theta)-\cos(\theta+\epsilon)$ on $[0,\frac{\pi}{2})$.

Observing the Eq.~(\ref{eq:error_bound_cosine}), we realize the quantization errors for different intervals are not fixed. Since $\sin(\cdot)$ is monotonically increasing on $[b,\frac{\pi}{2})$, the quantization errors $|v_1 - Q_v(v_1)| < |v_2 - Q_v(v_2)|$ if $|v_1| > |v_2|$. As a result, our method quantizes the gradients with larger magnitudes more precisely, which is important to training~\cite{ruder2016overview_sgd,lin2018deep_gradient_compression}.

Further, we then compare the quantization errors between the linear method and ours.
The biased linear quantization has an error bound of $\frac{b_v}{2^s-1}=\frac{\cos(b_{\theta})}{2^s-1}\cdot\|\textbf{v}\|_2$ for all the gradients on $[-b_v,b_v]$. Therefore, 
for the $k$-th quantization interval, our method has smaller errors than the linear if

\vspace{-0.2cm}
\begin{equation} \label{eq:error_condition}
\begin{split}
  2\cdot\sin(q\cdot(k+\frac{3}{4}))\cdot\sin(q\cdot\frac{1}{4}) < \frac{\cos(b_{\theta})}{2^s-1}.
\end{split}
\end{equation}

In the Eq.~(\ref{eq:error_condition}), only variable $k$ affects the result. As a result, top $66.6\%$, $40\%$ and $42.35\%$ quantization intervals from our 2-, 4- and 8-bits compression have smaller error bounds than the linear method. 
Interestingly, even though the quantization errors from our method are larger than the linear for most variables in a gradient vector, we are
able to show more favorable results with higher model performance in section~\ref{sec:exp}.
This observation and analysis are inspiring to help us understand gradient quantization in that the success of recovering larger gradients precisely plays a critical role in training.

Besides, our quantization also has an opposite observation to many previous work~\cite{han2015deep_compression,zhang2018lqnet,fu2020tinyscript}, which learns finer quantization intervals for densely distributed ranges. However, the large gradients are rarely distributed, as an example shown in Fig.~
\ref{fig:teaser}.
In contrast, our solution shows the importance of gradients is also critical for preserving high performance, instead of their distribution only.

\subsection{Complexity analysis}
\label{subsec:complexity}
We analyze and compare the complexity of our quantization and previous methods~\cite{alistarh2017qsgd,konevcny2016fl_communication,suresh2017distributed_mean_hadamard,han2015deep_compression,fu2020tinyscript}. Table~\ref{tab:time_complexity} reports the computation complexity of quantization and dequantization. Let us consider a $n$-bits quantization for a $m$-d vector, which leads to a $2^n$ level compression.
We show that the linear quantization~\cite{alistarh2017qsgd} and our method have the lowest complexity at $\mathcal{O}(m)$ because both methods have a closed-form quantization. Further, despite the improved performance from random Hadamard rotations~\cite{konevcny2016fl_communication,suresh2017distributed_mean_hadamard}, it increases the complexity of compression due to the matrix-vector multiplication. Regarding nonlinear quantization, $k$-means based methods~\cite{han2015deep_compression} have $\mathcal{O}(m^2)$ for clustering and $\mathcal{O}(mn)$ for searching. Last, TinyScript~\cite{fu2020tinyscript} also searches on irregular intervals for each dimension at $\mathcal{O}(\log 2^n)=\mathcal{O}(n)$. Consequently, we can clearly see the advantages of our method in terms of computation complexity since our approach needs less computation than other nonlinear approaches, making our method in particular suitable to deploy on edge devices with limited resources.

Additionally, we also discuss the memory costs generated for quantization and dequantization. First of all, we emphasize that our method only needs two float numbers (i.e., bound $b_{\theta}$ and norm $\|\textbf{v}\|_2$ ) to perform dequantization. Therefore, the additional communication costs for our method is negligible compared to quantized gradients. Even though the linear quantization and its improved version with random Hadamard rotations also need only two float numbers for defining the quantization range, we point out their performances remain improvable in section~\ref{sec:exp}. On the other hand, $k$-means needs additional costs of $\mathcal{O}(m)$ for the clustering centroids, leading to more data to communicate between a server and clients. In the end, TinyScript has to create quantization tables for each client, additionally introducing memory loading for edge devices.
To conclude, our proposed method has a simple closed-form computation and does not need a lot of additional memory costs, which is suitable for federated learning.

\begin{table}[t]
\begin{minipage}[t]{0.52\textwidth}

\scriptsize
\centering
\caption{Comparison of the computational complexity for $m$-d vectors using $n$-bits quantization, including linear (L)~\cite{alistarh2017qsgd}, linear with random Hadamard rotations (L+R)~\cite{konevcny2016fl_communication,suresh2017distributed_mean_hadamard}, $k$-means (KM)~\cite{han2015deep_compression}, TinyScript (TS)~\cite{fu2020tinyscript} and ours. }
\begin{tabular}{p{0.06\textwidth}<{\centering}p{0.14\textwidth}<{\centering}lll}
 \toprule
      L & L+R & KM & TS  & Ours \\
    \cmidrule(lr){1-5}
    $\mathcal{O}(m)$  & $\mathcal{O}(m+m\log m)$ & $\mathcal{O}(m(m+n))$ & $\mathcal{O}(mn)$ & $\mathcal{O}(m)$ \\
    \bottomrule
\end{tabular}
\vspace{-0.04cm}

\label{tab:time_complexity}

\end{minipage}
\begin{minipage}[t]{0.03\textwidth}
\quad
\end{minipage}
\begin{minipage}[t]{0.45\textwidth}

\scriptsize
\centering
\caption{Comparison results ($\%$) to previous quantization methods on CIFAR-10 ($B=50, E=5, C=0.1$). Different bits are applied.}
\begin{tabular}{l|p{0.03\textwidth}<{\centering}p{0.03\textwidth}<{\centering}p{0.03\textwidth}<{\centering}|p{0.03\textwidth}<{\centering}p{0.03\textwidth}<{\centering}p{0.03\textwidth}<{\centering}}
 \toprule
      \multirow{2}{*}{Method}  & \multicolumn{3}{c}{Biased ($n$-bits) } & \multicolumn{3}{c}{Unbiased ($n$-bits)} \\
       &  8 & 4 & 2 & 8 & 4 & 2 \\
    \cmidrule(lr){1-7}
    float32  &  \multicolumn{6}{c}{85.2 }\\
    linear &  85.18 & 85.15 & 10 & 85.19 & 85.16 & 73.11 \\
    $k$-means & 85.17 & 85.15 & 82.1 & - & - & - \\
    Ours &  85.3 & 85.24 & 85.17 & 85.28 & 85.25 & 85.2 \\
    \bottomrule
\end{tabular}
\vspace{-0.04cm}

\label{tab:cifar10}

\end{minipage}
\end{table}

\vspace{-0.1cm}
\subsection{Overall system}
\label{subsec:overall_system}
Despite the fruitful merits, we further reduce the costs with a few techniques. First, \textit{Deflate} algorithm~\cite{deutsch1996deflate} is a lossless data-dependent compression method and widely used in applications such as gzip files. We apply the \textit{Deflate} to compress the quantized messages in this work. Second, our method can employ parallel work for further cost reduction. In this work, we utilize random sparsification~\cite{konevcny2016fl_communication}, a common technique for gradient sparsification, to send parts of the gradients and greatly save the communication cost while maintaining comparable performance. As a result, we reduce the server-to-client and client-to-server communication cost significantly.

\vspace{-0.1cm}
\section{Experiments}
\vspace{-0.2cm}
\label{sec:exp}

We conduct our experiments of federated learning using FedAvg~\cite{mcmahan2017fl_google} on image classification with CIFAR-10 and brain tumor semantic segmentation with BraTS2018 dataset. All the methods are implemented with PyTorch~\cite{paszke2019pytorch} and PySyft~\cite{ryffel2018pysyft}. To test the broad optimization scenarios, we employ SGD~\cite{robbins1951sgd,sutskever2013sgd_momentum} or Adam~\cite{kingma2015adam} as the basic learners for the clients. Besides, we train the baseline models with float32 and FedAvg, followed by gradients quantization with 8-, 4- and 2-bits compression. For all the settings with quantization (i.e., the baselines and ours), we utilize layer-wise quantization on the neural networks. By default, we apply the biased quantization for ours and perform gradient clipping on top $1\%$ gradients to obtain the bound $b_{\theta}$ as depicted in section~\ref{sec:method}.

We compare our nonlinear quantization with the standard linear-based method, linear-based probabilistic unbiased quantization (denoted by linear (U))~\cite{alistarh2017qsgd}, as well as the improved version (denoted by linear (U, R))~\cite{konevcny2016fl_communication} with random Hadamard rotations~\cite{suresh2017distributed_mean_hadamard}. Further, we also compare our method with previous nonlinear methods $k$-means based approach~\cite{han2015deep_compression} and recent proposed TinyScript~\cite{fu2020tinyscript}. Finally, we compare our method with 1-bit compression, including signSGD~\cite{bernstein2018signsgd}, and its improved version with error feedback (+EF)~\cite{karimireddy2019error_feeback_signsgd} and error assimilation (+CSEA)~\cite{xie2020cser}. In our experiment, we first compare the gradient compression in section~\ref{subsec:comp_gradients}, and then discuss the round-trip cost reduction.

\subsection{Datasets and setup}
\vspace{-5pt}

We conduct experiments using CIFAR-10 dataset for image classification, following~\cite{mcmahan2017fl_google}. We apply a CNN model~\cite{tf_tutorial} with 122,570 parameters. For FedAvg, we set $E=5$ and $B=50$, and $C=0.1$.

Besides, we evaluate on 3D semantic segmentation using BraTS dataset, following~\cite{sheller2018multi_brain_fl}. We adopt a 3D-UNet~\cite{cciccek20163d_unet} with 9,451,567 parameters for this task.
For FedAvg, we set $C=1$, $E=3$, and $B=3$. Each worker adopts Adam with betas equal to (0.9, 0.999). BraTS 2019 contains 335 training examples, which covers the BraTS 2018 training set. Therefore, we train a model on the BraTS2018 training set and compute the dice score on the 50 unseen examples.\footnote{BraTS datasets do not provide GT for the validation set.}

We train federated models with gradient compression only to compare the performance of different approach in quantizing gradients, as depicted in section~\ref{subsec:comp_gradients}. Further, we apply quantization on model weights and gradients to reduce the overall communication cost in section~\ref{subsec:comp_round_trip}.

\begin{figure}[!t]
\begin{minipage}[t]{0.48\textwidth}
\begin{center}
\includegraphics[width=0.445\linewidth]{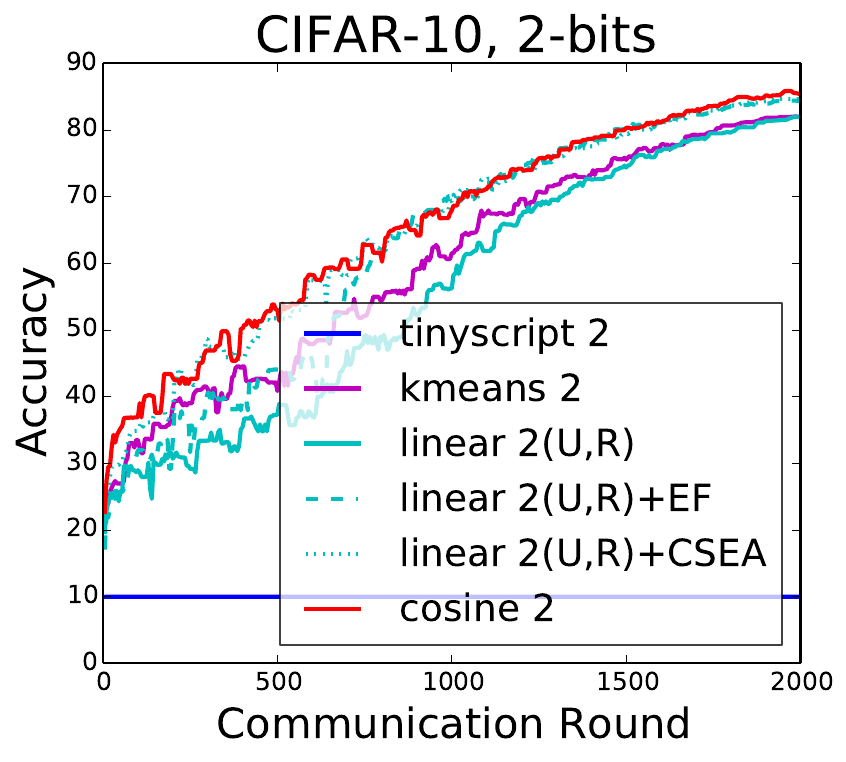}
\includegraphics[width=0.445\linewidth]{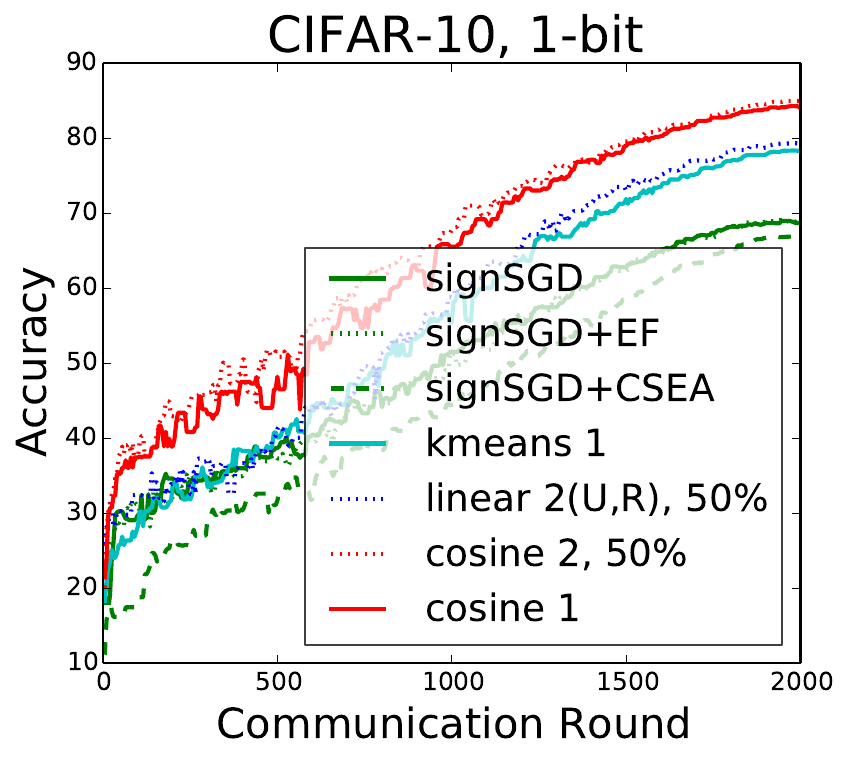}
   \end{center}
   \vspace{-0.4cm}
   \caption{Low-bits comparison results ($\%$) to various compression schemes on CIFAR-10 ($B=50, E=5, C=0.1$). }
    \vspace{-0.35cm}
\label{fig:cifar10_low_bits}
\end{minipage}
\begin{minipage}[t]{0.04\textwidth}
\end{minipage}
\begin{minipage}[t]{0.48\textwidth}
\begin{center}
\includegraphics[height=0.405\linewidth,width=0.46\linewidth]{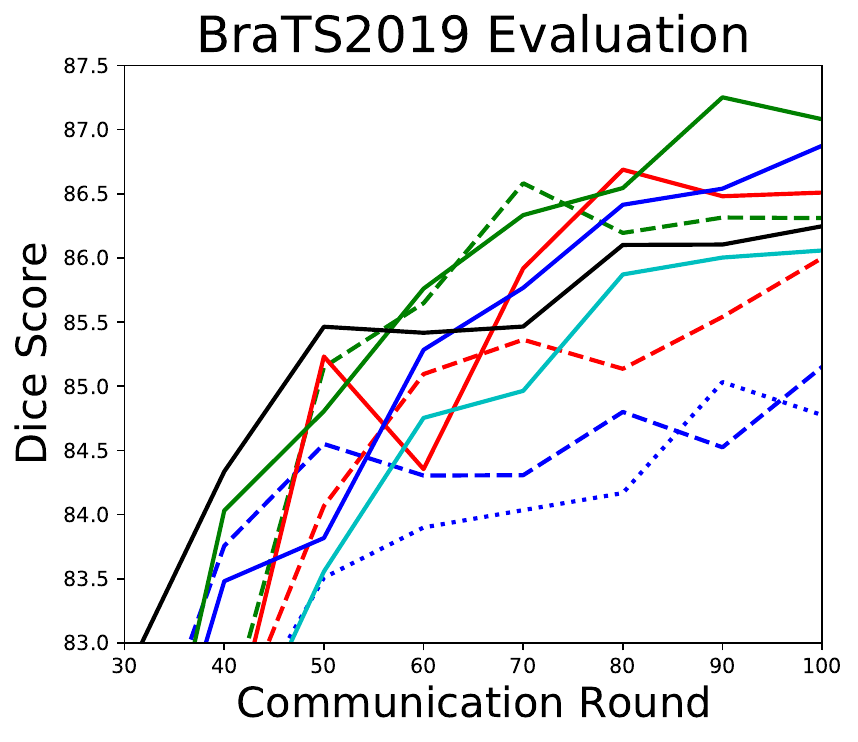}
\includegraphics[width=0.433\linewidth]{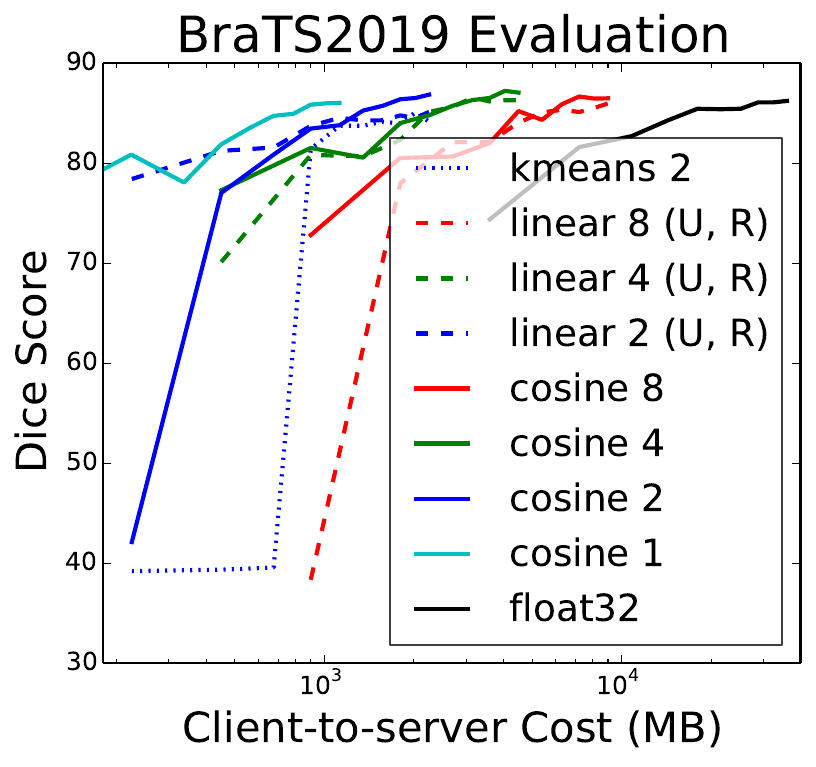}
   \end{center}
   \vspace{-0.4cm}
   \caption{Comparison results ($\%$) to other methods on BraTS2019 dataset ($B=3, E=3, C=1$).  }
\vspace{-0.4cm}
\label{fig:brats_results}
\end{minipage}

\end{figure}

\subsection{Experimental results}
\vspace{-8pt}
\label{subsec:comp_gradients}

\begin{table}[h!]

\begin{minipage}[h]{0.5\textwidth}
\myparagraph{Full gradient compression} We compare our quantization with linear quantization~\cite{alistarh2017qsgd} and $k$-means based method~\cite{han2015deep_compression} on CIFAR-10,
as listed in Table~\ref{tab:cifar10}. Although unbiased quantization (e.g. Eq.~\ref{eq:unbiased_quan}) indeed improves the performance, biased quantization is still helpful for understanding the effect of quantization intervals. Therefore, we conduct experiments using both biased and unbiased version for linear quantization~\cite{alistarh2017qsgd} and our methods. 

\end{minipage}
\qquad
\begin{minipage}[h]{0.45\textwidth}

\scriptsize
\centering
\caption{Comparison results ($\%$) to previous quantization methods on CIFAR-10 ($B=50, E=5, C=0.1$). Different bits are applied.}
\begin{tabular}{l|p{0.04\textwidth}<{\centering}p{0.04\textwidth}<{\centering}p{0.04\textwidth}<{\centering}|p{0.04\textwidth}<{\centering}p{0.04\textwidth}<{\centering}p{0.04\textwidth}<{\centering}}
 \toprule
      \multirow{2}{*}{Method}  & \multicolumn{3}{c}{Biased ($n$-bits) } & \multicolumn{3}{c}{Unbiased ($n$-bits)} \\
       &  8 & 4 & 2 & 8 & 4 & 2 \\
    \cmidrule(lr){1-7}
    float32  &  \multicolumn{6}{c}{85.2 }\\
    linear &  85.18 & 85.15 & 10 & 85.19 & 85.16 & 73.11 \\
    $k$-means & 85.17 & 85.15 & 82.1 & - & - & - \\
    Ours &  85.3 & 85.24 & 85.2 & 85.28 & 85.25 & 85.2 \\
    \bottomrule
\end{tabular}

\label{tab:cifar10}

\end{minipage}
\vspace{-10pt}
\end{table}

We highlight the following observations: (1) Even though  all the methods work well for high-bits compression, it remains challenging to maintain high model performance in the low-bits case, where only our 2-bits method can achieve comparable results to float32 based training. In particular, 2-bits linear quantization based training has a clear gap to the float based training, even unbiased version is applied. (2) Even though the $k$-means based method also adopts a non-linear interval, it fails to reach the performance of float32-based training (i.e., $82.1\%$ vs. $85.2\%$ in CIFAR-10), whereas our method resembles the performance of full precision. It clearly supports our hypothesis that larger gradients play a critical role in training.

From the above results, the low-bits cases are the key to achieve breakthroughs. Hence, we focus on more challenging 2-bits and 1-bit compression and compare our methods to recent advanced methods including~\cite{konevcny2016fl_communication,karimireddy2019error_feeback_signsgd,bernstein2018signsgd,xie2020cser,fu2020tinyscript}. 
Fig.~\ref{fig:cifar10_low_bits} plots the performance curves. From our previous results, we show 2-bits unbiased linear quantization is not enough to achieve decent results, therefore, we boost it with random Hadamard rotations~\cite{suresh2017distributed_mean_hadamard,konevcny2016fl_communication}, namely linear 2(U,R). We also apply the error feedback (EF)~\cite{karimireddy2019error_feeback_signsgd} and error assimilation (CSEA)~\cite{xie2020cser} on linear 2(U,R) and signSGD. Last, we also combine $50\%$ random sparsification~\cite{konevcny2016fl_communication} for linear 2(U,R) and cosine 2, resulting in comparable communication costs for gradients. From this plot, we emphasize several aspects: (1) Our quantization performs quite well in all the cases. Our 1-bit compression only decreases the performance of float32 a little in CIFAR-10 (i.e., $84.2\%$ vs. $85.2\%$). (2) Even though the EF and CSEA improves 2-bits linear quantization, it is still worse than our method in performance and it needs to store the residual gradients on each client. Further, the improvement of those two methods is not significant in 1-bit compression due to less frequent interactions in federated learning. (3) We achieve better results than $k$-means~\cite{han2015deep_compression,fu2020tinyscript}, which obviously demonstrates the effectiveness of our quantization.

In addition to image classification, we show the results for brain tumor semantic segmentation in Fig.~\ref{fig:brats_results} using BraTS2019 dataset. Because of the effectiveness of unbiasedness and Hadamard rotations~\cite{konevcny2016fl_communication} from our previous comparison, we only apply the unbiased linear quantization with  Hadamard rotations for BraTS2019, which is the strongest setup for the linear. Besides, $k$-means based method is very expensive for high-bits quantization in the 3D-UNet for segmentation, therefore, we only apply 2-bits $k$-means for comparison. From Fig.~\ref{fig:brats_results}, we also achieve better results (Dice score is applied) in this task. Besides, we measure the client-to-server cost over dice scores, which further validates the effectiveness of compression schemes, as drawn in the right plot of Fig.~\ref{fig:brats_results}. Clearly, we observe that our method achieves better performance while requiring much fewer costs.

\begin{figure}[!t]
\begin{center}
\begin{tabular}{@{}c@{}c@{}c@{}c}
\includegraphics[width=0.22\linewidth]{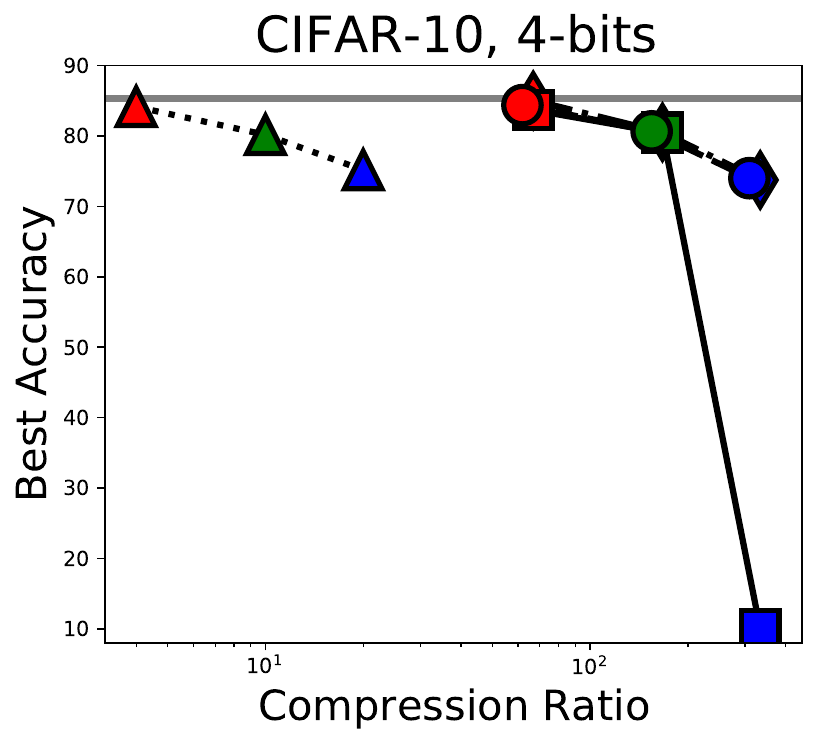} &
\includegraphics[width=0.22\linewidth]{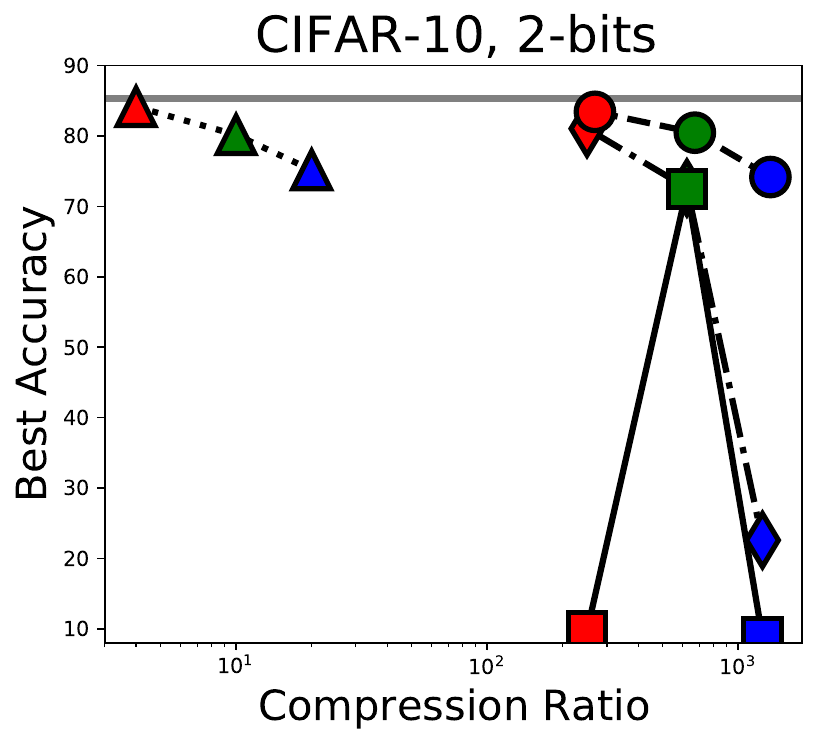} &
\includegraphics[width=0.215\linewidth]{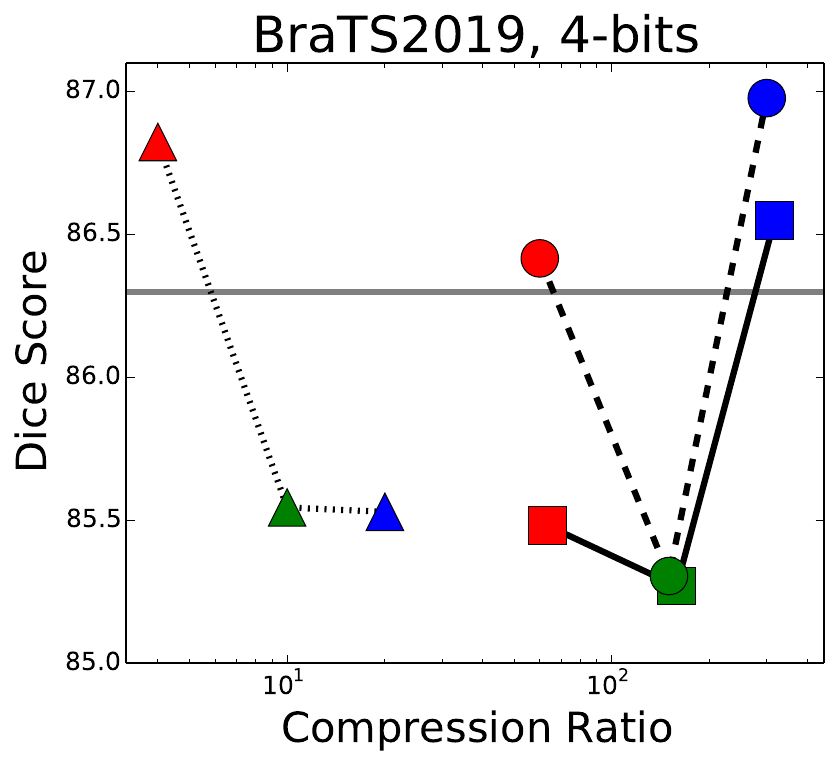} &
\includegraphics[width=0.215\linewidth]{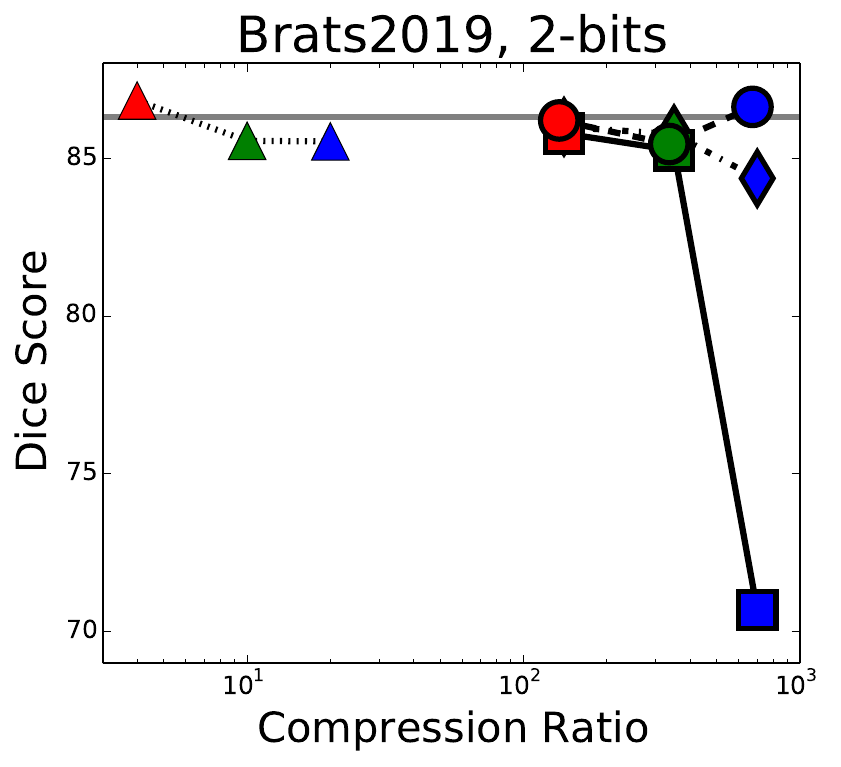} \\

\multicolumn{2}{c}{(a) CIFAR-10} & \multicolumn{2}{c}{(b) BraTS2019} \\

\multicolumn{4}{c}{ \includegraphics[width=0.45\linewidth]{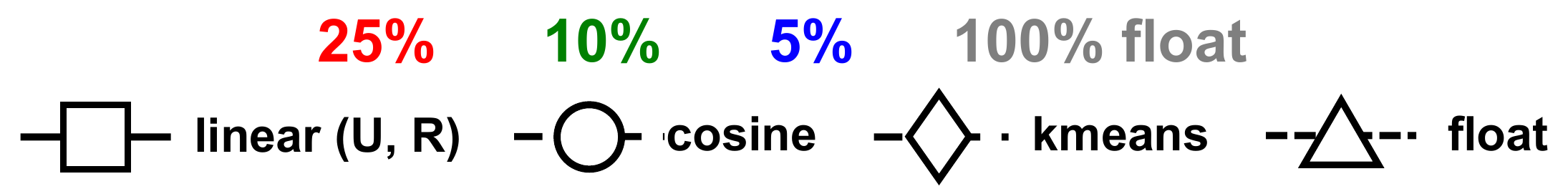} } \\
\end{tabular}
   \end{center}
   \vspace{-0.5cm}
   \caption{Comparison results ($\%$) on the combination of random sparsification~\cite{konevcny2016fl_communication}. The $x$-axis with the logarithmic scale indicates the compression ratios on the gradients. }
\label{fig:cifar10_compressed_random}
\vspace{-0.7cm}
\end{figure}

\myparagraph{Combination with gradient sparsification}
We report the results by combining gradient sparsification, {\it Deflate} compression  and different quantization methods (2- \& 4-bits are considered), as shown in Fig.~\ref{fig:cifar10_compressed_random}. We return $25\%$, $10\%$ and $5\%$ gradients for each client to aggregate the global model.  In Fig.~\ref{fig:cifar10_compressed_random}, we first observe the unsuitability of linear, leading to fluctuating performance. Second, we release $k$-means fails to maintain the performance when 2-bits and 5\% sparsification are employed. In contrast, our method can still train a model with good performance and reach very high compression ratio, such as more than 1000$\times$ when the extreme compression is applied.

\begin{figure}[t]
\vspace{-25pt}
\begin{center}
\includegraphics[width=0.45\linewidth]{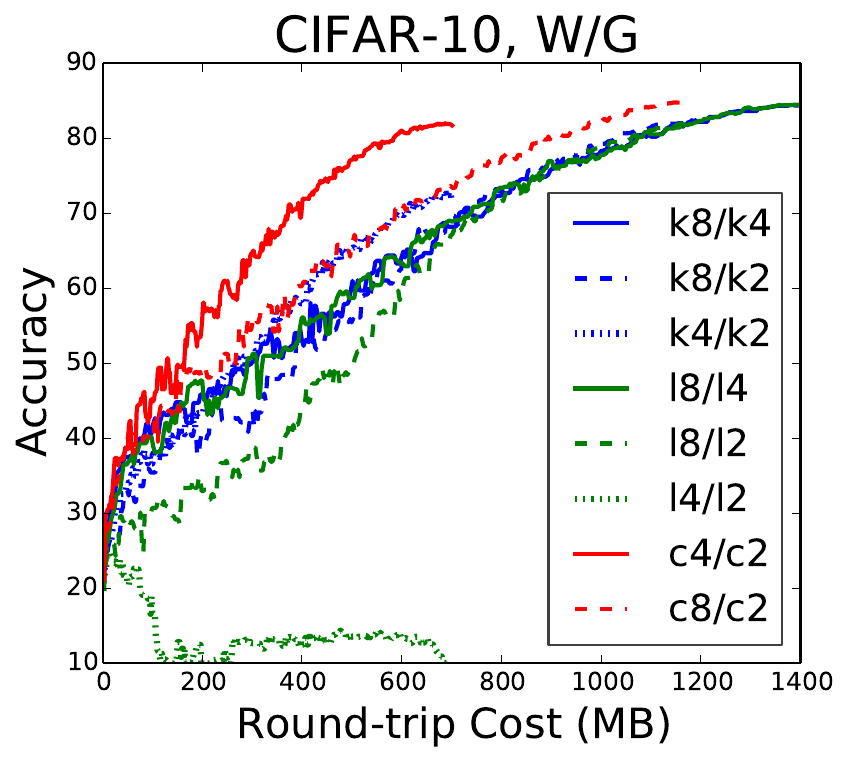}
   \end{center}
   \vspace{-0.5cm}
   \caption{Comparison of double quantization on CIFAR-10.  k: $k$-means; l: linear (U,R); c: cosine. W: weights; G: gradients.}
   \vspace{-0.5cm}
\label{fig:cifar10_model_weights}
\end{figure}

\myparagraph{Reducing overall costs}
\label{subsec:comp_round_trip}
In the end, we employ and compare different quantization methods for compressing model weights and gradients. First, we compress the model weights using different methods with various precision. Second, we apply different combincation of various quantization for jointly compressing server-to-client (weights) and client-to-server (gradients) costs. All the results are drawn in Fig.~\ref{fig:cifar10_model_weights}. In this experiment, only unbiased linear quantization with Hadamard rotations is tested, because it gives the best performance compared to other versions of linear quantization.
We clearly observe the advances of our method. In terms of  compressing model weights, our quantization still works well, and achieves better performance over the same level of communication costs. Besides, we can see that only our method is able to compress model weights with 4-bits quantization.

\vspace{-0.2cm}
\section{Conclusion}
\vspace{-0.3cm}
We propose a simple quantization scheme, which is in principle different from previous methods - we consider the effects of values to be quantized instead of their distribution. Inspired by the observation that gradients/weights with large magnitudes are more informative, we quantize values according to a non-uniform interval induced by cosine functions, in which larger values have finer levels while the others have coarser levels. Due to the simplicity of cosine functions, our method is a favorable choice for federated learning.
We successfully apply the proposed method on model weights and gradients to reduce the round-trip communication cost in federated learning. 
Furthermore, by incorporating gradient sparsification and lossless data compression, extensive results show that our method is more effective than previous widely used linear and non-linear quantization to reduce communication costs and train a model with simply computation steps at decent performance.

\section*{Acknowledgement}
This work is partially funded by the Helmholtz Association within the project ”Trustworthy Federated Data Analytics (TFDA)” (ZT-I-OO1 4).

\bibliography{egbib}
\bibliographystyle{abbrvnat}

\end{document}